\documentclass[11pt]{article}
\usepackage{paclic32}
\usepackage{times}
\usepackage{latexsym}
\usepackage{amsmath}
\usepackage{multirow}
\usepackage{url}
\usepackage[round]{natbib}

\usepackage{graphicx, xcolor}
\usepackage{latexsym}
\usepackage{booktabs}
\usepackage{amsmath,amssymb}
\usepackage{tikz}
\usepackage{multirow,bigdelim}
\newcommand{\FFT}{\mathop{\mathfrak{F}}}

\def\Underline{\setbox0\hbox\bgroup\let\\\endUnderline}
\def\endUnderline{\vphantom{y}\egroup\smash{\underline{\box0}}\\}
\def\|{\verb\|}

\newcommand{\Cset}{\mathbb{C}}
\newcommand{\Rset}{\mathbb{R}}
\newcommand{\mat}[1]{\boldsymbol{\mathbf{#1}}}

\newcommand{\transpose}{^{\mathrm{T}}}

\newcommand{\diag}{\mathop{\text{diag}}}

\title{Reduction of Parameter Redundancy in Biaffine Classifiers \\ with Symmetric and Circulant Weight Matrices}

\author{Tomoki Matsuno$^{1}$, Katsuhiko Hayashi$^{2, 4}$, Takahiro Ishihara$^{1}$, Hitoshi Manabe$^{3}$, Yuji Matsumoto$^{1, 4}$ \\
  $^{1}$Nara Institute of Science and Technology \\
  $^{2}$Osaka University  $^{3}$Works Applications \\
  $^4$RIKEN Center for Advanced Intelligence Project\\
  \tt $^{1}$\{matsuno.tomoki.mr1,ishihara.takahiro.in0, matsu\}@is.naist.jp\\
  \tt $^{2}$khayashi0201@gmail.com $^{3}$manabe\_h@worksap.co.jp\\
}

\date{}

\begin{document}
\maketitle
\begin{abstract}
  Currently, the biaffine classifier has been attracting attention as a method to introduce an attention mechanism into the modeling of binary relations. For instance, in the field of dependency parsing, the Deep Biaffine Parser by \citeauthor{deepbiaffine} has achieved state-of-the-art performance as a graph-based dependency parser on the English Penn Treebank and CoNLL 2017 shared task. On the other hand, it is reported that parameter redundancy in the weight matrix in biaffine classifiers, which has $O(n^2)$ parameters, results in overfitting ($n$ is the number of dimensions). In this paper, we attempted to reduce the parameter redundancy by assuming either symmetry or circularity of weight matrices. In our experiments on the CoNLL 2017 shared task dataset, our model achieved better or comparable accuracy on most of the treebanks with more than 16\% parameter reduction.
\end{abstract}

\section{Introduction}

  Recently, methods based on attention mechanisms have been used in various fields of natural language processing \citep{attention, attention-summarization}. In tasks which deal with \textbf{binary relations} that take two words as arguments, e.g., dependency parsing, a good number of these models have achieved high performance \citep{mlp-attention, bilinear-attention, deepbiaffine}.
  
  \textbf{Biaffine transformation} is a method to incorporate an attention mechanism into binary relations proposed by \cite{deepbiaffine} (following them, we call this method \textbf{biaffine classifier}). They achieved the state-of-the-art performance among graph-based dependency parsers for the English Penn Treebank. In addition, the state-of-the-art transition-based parser on English Penn Treebank uses a biaffine classifier to evaluate the probability distribution of a word coming into the stack at a step point~\citep{stackpointer}.
  
  While biaffine transformation has rich expressiveness in modeling binary relations, its number of parameters in the weight matrix (bilinear term) is $O(n^2)$ (where $n$ is the number of dimensions). This redundant parameters can result in high degree of freedom of the model, thus causing overfitting especially when a large number of training samples are not available~\citep{hole}.  
  
  In this paper, we attempt to reduce the redundancy by introducing the assumption of either symmetry or circularity in the weight matrix at a biaffine classifier. With either assumption, we can vectorize the matrix and reduce the space complexity to $O(n)$. Additionally, the time complexity becomes $O(n)$ in the case of symmetry and $O(n\log n)$ in the case of circularity with the fast Fourier transform. 
Furthermore, while the expressiveness of the model based on symmetry is restricted\footnote{Denote the score of the bilinear term by $f(a,b)$ when a word pair $a$, $b$ has a dependency relation. In this case, if the bilinear term has symmetry, $f(a,b)=f(b,a)$ . Therefore, this is not appropriate for expression of a directed edge. },  one based on circularity is able to express asymmetry relations.

In our experiments, we imposed constraints on the biaffine classifiers of the deep biaffine parser and examined the effect on the accuracy of the model. For our experiments, we used the dependency parsing dataset from the CoNLL 2017 shared task~\citep{conll2017}. We chose four languages which have relatively rich training examples and four languages which have fewer. From our experiments, we found that the method with the circularity assumption outperformed the baseline in most of the languages.

\section{Deep Biaffine Parser}

Models introduced in this paper are based on the \textbf{Deep Biaffine Parser} proposed by \citep{deepbiaffine}. They achieved the state-of-the-art accuracy on the CoNLL 2017 shared task for Universal Dependencies~\citep{conll2017}. 

This model receives a sequence of words and POS tags, and calculates the probability of an arc between each pair of words as well as a syntactic function label for each arc. For evaluation of scores, it uses Long Short-Term Memories (LSTMs), Multi-Layer Perceptrons (MLPs) and biaffine classifiers.

In the following sections, first, we explain the biaffine transformation which is the essential part of a biaffine classifier while skipping explanation about LSTM and MLP for the sake of simplicity. Then, we describe the overview of the model. It is worth noting that the structure of the model is different from that in \citep{deepbiaffine} in that it utilizes character level information. This is because we used an updated version of the model that was made for the shared task \citep{dozat2017stanford}. 

\subsection{Biaffine Transformation}

For the dependency parsing score functions, we use the biaffine transformation shown below to model binary relations. Here, $\oplus$ stands for concatenation of vectors. The first term on the right side represents relatedness score, and the second term the score of $v_i$ and $v_j$ appearing independently.  $b$ is bias term.

\begin{equation}\label{eq:biaffine}
\begin{split}
g(\mathbf{v}_i, \mathbf{v}_j) 
&= 
\mathbf{v}_i\transpose\mathbf{Av}_j + (\mathbf{v}_i \oplus \mathbf{v}_j) \transpose \mathbf{b} + b
\end{split}
\end{equation}

\subsection{Structure of the Model}

We show the structure of this model below.

\begin{enumerate}

\item
First, this model takes two sequences: words and POS tags. It uses a unidirectional LSTM to encode each words' character-level information into a vector. It then sums this vector with a separate token-level word embeddings. 

 \item
It then concatenates the vectors obtained above with POS embeddings and encodes them via a three layer bidirectional LSTM. $y_i$ denotes a vector made by concatenating the hidden states (not including cell states) from the LSTMs of both directions which corresponds to the $i$ th word $w_i$.

\item
These outputs are then transformed with MLPs. Here, we use distinct MLPs for dependents and heads in the prediction of arcs and labels.
\begin{equation}
\begin{split}
\mathbf{v}_{i}^{arc-head} &= \mathrm{MLP}^{arc-head}(\mathbf{y}_i),\\
\mathbf{v}_j^{arc-dep} &= \mathrm{MLP}^{arc-dep}(\mathbf{y}_j), \\
\mathbf{v}_{i}^{label-head} &= \mathrm{MLP}^{label-head}(\mathbf{y}_i),\\
\mathbf{v}_j^{label-dep} &= \mathrm{MLP}^{label-dep}(\mathbf{y}_j).
\end{split}
\end{equation}

where $\mathbf{v}_{i}^{arc-head}, \mathbf{v}_{j}^{arc-dep} \in \mathbb{R}^{n}$ \\
and $\mathbf{v}_{i}^{label-head}, \mathbf{v}_{j}^{label-dep} \in \mathbb{R}^{m}$ \\

\item
The scores of arcs between each word pair are calculated using a biaffine transformation.
\begin{equation} \label{eq:arc}
s_{i,j}^{(arc)} = {\mathbf{v}_i^{arc-head}}^{\mathrm T} \mathbf W \mathbf{v}_j^{arc-dep} + {\mathbf{v}_i^{arc-head}}^{\mathrm T} \mathbf b^{(arc)}.
\end{equation}

\item
Running Chu-Liu/Edmonds algorithm \citep{mst, edmonds} on the scores calculated in (3), we obtain a tree structure that maximizes the total score.

\item
It evaluates a score $s_{i,j}^{(l)}$  of assigning a label $l$ $(l \in \{1, 2, ..., L\};$ $L$: the number of labels) on the arc between the dependent word $w_i$ and its predicted head word $w_j$. The equation is defined below.

\begin{equation} \label{eq:label}
\begin{split}
s_{i,j}^{(l)} 
&= {\mathbf{v}_{i}^{label-head}}\transpose \mathbf{U}^{[l]} {\mathbf v_j^{label-dep}} \\
&+ {({\mathbf{v}_{i}^{label-head}} \oplus \mathbf v_j^{label-dep})}\transpose \mathbf{b}^{[l]} \\
&+ b^{[l]}.
\end{split}
\end{equation}

Here, a distinct weight matrix $\mathbf{U}^{[l]}$, weight vector $\mathbf{b}^{[l]}$ and bias $b^{[l]}$ are used for each label. The first term on the right side of the equation (4) represents the score of assigning the label $l$ to the arc between dependent $w_i$ and head $w_j$. The second term expresses the score of the label when the dependent and head are given independently.

\end{enumerate}

In our experiment, $\mat{W}\in\Rset^{n\times n}$ and $\mat{U}^{[l]}\in\Rset^{m\times m}$ account for about 17\% of the parameters in the deep biaffine parser. By reducing these parameters, we can expect not only improved memory efficiency but also less overfitting.

\section{Proposed Methods}
\label{sec:proposed}
In this section, we introduce our two proposed methods to reduce the number of parameters of the model. We impose either a symmetry or circularity constraint on the weight matrix $\mathbf{W}$ of (\ref{eq:arc}) and $\mathbf{U}^{[l]}~(\forall l\in\{1,2,\dots,L\})$ of (\ref{eq:label}).

\subsection{Symmetric Matrix Constraint}
This method assumes that the weight matrices of the bilinear terms are \textbf{symmetric matrices} and thus are diagonalizable. As a result, we can transform the bilinear term of the score functions into a ``triple inner product'' of two input vectors and a weight vector.
\subsubsection{Diagonalization of the weight matrix in the bilinear term}
When a matrix $\mathbf{W} \in \mathbb{R}^{n \times n}$ is symmetric, it can be diagonalized by an orthogonal matrix $\mathbf{O} \in \mathbb{R}^{n \times n}$ as below:
\[
\mathbf{W} = \mathbf{O} {\diag(\mathbf{w})} \mathbf{O}\transpose
\]
where $\mathbf{w} \in \mathbb{R}^{n}$ consists of the eigenvalues of $\mathbf{W}$ and $\diag(\mathbf{w}) \in \mathbb{R}^{n \times n}$ represents the diagonal matrix whose diagonal elements are $\mathbf{w}$. With this property, we can rewrite the bilinear term as follows:
\begin{equation}
\begin{split}
\label{eq:diag}
\mathbf{v}_i\transpose \mathbf{W} \mathbf{v}_j
&=
\mathbf{v}_i\transpose \mathbf{O} \diag(\mathbf{w}) \mathbf{O} \transpose \mathbf{v}_j  \\ &=
{\mathbf{v}^{\prime}_{i}}\transpose \diag(\mathbf{w}) \mathbf{v}^{\prime}_{j} \\
&=
\langle\mathbf{v}^{\prime}_{i}, \mathbf{w}, \mathbf{v}^{\prime}_{j}\rangle.
\end{split}
\end{equation}
where, $\mathbf{v}^{\prime}_{i}=\mathbf{O}\transpose \mathbf{v}_i$ and $\mathbf{v}^{\prime}_{j} = \mathbf{O}\transpose \mathbf{v}_j$, assuming $\mat{O}$ is learned implicitly. 
$\langle\mathbf{v}^{\prime}_{i}, \mathbf{w}, \mathbf{v}^{\prime}_{j}\rangle$ is a ``triple inner product'' of $\mathbf{v}^{\prime}_{i}, \mathbf{w}$ and $\mathbf{v}^{\prime}_{j}$ defined by $\langle \mathbf{a}, \mathbf{b}, \mathbf{c} \rangle = \sum_{k=1}^n a_k b_k c_k$. Consequently, the symmetry constraint on the matrix can reduce the number of weight parameters from $n^2$ to $n$.

\subsubsection{Simultaneous Diagonalization}
\label{sim-diag}
When a set of symmetric matrices forms a commuting family, they can be diagonalized by the same orthogonal matrix~\citep{liu2017analogy}. 
So we assume the weight matrices $\mathbf{U}^{[1]}, \mathbf{U}^{[2]}, \dots, \mathbf{U}^{[L]}$ of the scoring functions (\ref{eq:label}) form a commuting family. Namely, we assume:
\[
\mathbf{U}^{[p]}\mathbf{U}^{[q]} = \mathbf{U}^{[q]}\mathbf{U}^{[p]},~\forall p, q \in \{1,2, \dots, L\}.
\]
With this assumption, all $L$ weight matrices can be diagonalized simultaneously. Therefore, the vectors $\mathbf{v}_{i}^{label-head}$ and $\mathbf{v}_{j}^{label-dep}$ can be mapped by the same orthogonal matrix for all score functions.
\subsubsection{Score Functions}
Based on the above, under the assumption that all weight matrices are symmetric, we substitute the bilinear term in the biaffine transformations with a triple inner product. First, the score function for arc is defined as follows:
\begin{equation}
\label{eq:1arc}
\begin{split}
s_{i,j}^{(arc)} &= \langle {\mathbf{v}_i^{arc-head}}, \mathbf{w}, \mathbf{v}_j^{arc-dep} \rangle \\
&+ ({\mathbf v_{i}^{arc-head} \oplus \mathbf v_j^{arc-dep}})^{\mathrm T} \mathbf{b}
.
\end{split}
\end{equation}
where, $\mathbf{w} \in \mathbb{R}^n,~\mathbf{b} \in \mathbb{R}^{2n}$. Unlike in (\ref{eq:arc}), the second term of this function contains the arc-dep vector because we confirm that it improves the performance.

Scoring functions for labels are defined as follows:
\begin{equation} \label{eq:1label}
\begin{split}
s_{i,j}^{(l)} &= \langle {\mathbf v_{i}^{label-head}}, \mathbf{u}^{[l]}, {\mathbf v_j^{label-dep}} \rangle  \\
&+ ({\mathbf v_{i}^{label-head} \oplus \mathbf v_j^{label-dep}})^{\mathrm T} \mathbf {b}^{[l]}
.
\end{split}
\end{equation}
where, $\mathbf{u}^{[l]} \in \mathbb{R}^{m}~,\mathbf{b}^{[l]} \in \mathbb{R}^{2m}$. We eliminate the bias term $b^{[l]}$ in (\ref{eq:label}) because we confirm that it does not affect the performance.

\subsection{Circulant Matrix Constraint}
\cite{hole} used a circulant matrix for the bilinear transformation in knowledge graph completion model~\citep{rescal} to reduce the number of parameters and improve the computational efficiency. Following this method, we assume the weight matrices of the bilinear term in the biaffine transformations are circulant and propose new scoring functions based on that.

\subsubsection{Bilinear Transformation Using a Circulant Matrix}
We define the circulant matrix $C(\mathbf{w}) \in \mathbb{R}^{n \times n}$ for a vector $\mathbf{w} \in \mathbb{R}^{n} $ as follows:
\begin{equation}
C(\mathbf{w})=
\begin{bmatrix}
    w_1       & w_n & \dots & w_3 & w_2 \\
    w_2       & w_1 & w_n &  & w_3 \\
    \vdots    & w_2 & w_1 & \ddots & \vdots \\
    w_{n-1}   &     & \ddots & \ddots & w_{n} \\
    w_{n}     &  w_{n-1} & \dots & w_{2} & w_{1}
\end{bmatrix}
.
\end{equation}
where, $\mathbf{w} \transpose = (w_1, \dots, w_n)$. . Then, we replace the bilinear term with one where the weight matrix is a circulant matrix $C(w)$ with $n$ parameters:

\begin{equation}
\label{eq:circ-bilinear}
\mathbf{v}_{i} \transpose C(\mathbf{w})  \mathbf{v}_{j}.
\end{equation}

\subsubsection{Score Functions}
We propose score functions that employ (\ref{eq:circ-bilinear}) as the bilinear term in the biaffine transformation. The score function for an arc is then defined as follows:

\begin{equation}
\label{eq:2arc}
\begin{split}
s_{i,j}^{(arc)} &= {\mathbf{v}_i^{arc-head}} C(\mathbf{w}) \mathbf{v}_j^{arc-dep}  \\
&+ ({\mathbf v_{i}^{arc-head} \oplus \mathbf v_j^{arc-dep}})^{\mathrm T} \mathbf{b}
.
\end{split}
\end{equation}
where, $\mathbf{w} \in \mathbb{R}^n,~\mathbf{b} \in \mathbb{R}^{2n}$.

The score functions for labels are defined as follows:
\begin{equation} \label{eq:2label}
\begin{split}
s_{i,j}^{(l)} &= {\mathbf v_{i}^{label-head}}C(\mathbf{u}^{[l]}){\mathbf v_j^{label-dep}} \\
&+ ({\mathbf v_{i}^{label-head} \oplus \mathbf v_j^{label-dep}})\transpose \mathbf{b}^{[l]}
.
\end{split}
\end{equation}
where, $\mathbf{u}^{[l]} \in \mathbb{R}^{m}~,\mathbf{b}^{[l]} \in \mathbb{R}^{2m}$.

\subsubsection{Efficient Computation Using Fast Fourier Transformation}
In this section, we explain how to compute (\ref{eq:circ-bilinear}) efficiently using a fast Fourier transformation (FFT). We denote an $n$-point discrete Fourier transformation (DFT) matrix as $\FFT _n\in \mathbb{C}^{n \times n}$. Then, any circulant matrix $C(\mathbf{w}) \in \mathbb{R}^{n \times n}$ can be diagonalized as follows~\citep{gray2006toeplitz}:

\[
C(\mathbf{w}) = {\FFT}_n^{-1}\diag({\FFT}_n \mathbf{w}){\FFT}_n.
\]

With this property, we can rewrite (\ref{eq:circ-bilinear}) as a triple hermitian inner product ~\citep{liu2017analogy}:
\begin{align}
\label{eq:prop2}
\mathbf{v}_{i} \transpose C(\mathbf{w}) \mathbf{v}_{j}
&= \mathbf{v}_{i} {\FFT}_n^{-1}\diag({\FFT}_n \mathbf{w}){\FFT}_n \mathbf{v}_{j} \nonumber\\
&= \frac{1}{n} {\overline{{\FFT}_n \mathbf{v}_{i}}}\transpose \diag({\FFT}_n \mathbf{w}){\FFT}_n \mathbf{v}_{j} \nonumber\\
&= \langle \mathbf{v}^{\prime}_{i}, \mathbf{w}^{\prime}, \mathbf{v}^{\prime}_{j} \rangle \nonumber\\
&= \Re (\langle \mathbf{v}^{\prime}_{i}, \mathbf{w}^{\prime}, \mathbf{v}^{\prime}_{j} \rangle).
\end{align}
Here, $\mathbf{v}^{\prime}_{i}=\overline{{\FFT}_n\mathbf{v}_i}$, $\mathbf{v}^{\prime}_j={{\FFT}_n\mathbf{v}_j}$,
${\mathbf{w}^{\prime}}=\frac{1}{n}\diag({\FFT}_n\mathbf{w})$, and all of them are $n$ dimensional complex vectors. $\Re(\cdot)$ is the operator which takes the real parts of its argument. With this transformation, we can compute the bilinear transformation with a circulant matrix in $O(n \log n)$ using a FFT.

The DFT of an $n$-dimensional vector $\mathbf{x}$,  ${\FFT}_n\mathbf{x}$, is conjugate symmetric if and only if $\mathbf{x}$ is a real vector~\citep{hayashi:17-2}.
In our experiment, we initialize $\mathbf{w}'$ with the DFT of a real vector and update it in complex space.  We update ``frequency'' domain (complex space) vectors using only the operations which have correspondence to ``time domain'' vectors. Thus, as described in \cite{hayashi:17-2}, the conjugate symmetry of vectors are kept while learning because their initial values satisfy it.

\subsubsection{Expressiveness of the Bilinear Transformation Using Circulant Matrices}
In this section, we explain about the expressiveness of circulant matrices in relation to an arbitrary matrix $\mat{W}\in\Rset^{n\times n}$. \cite{complex-journal} show that for any $\mat{W}$, there exists a normal matrix $\mat{W}'\in\Cset^{n\times n}$ such that $\mat{W}=\Re(\mat{W}')$. Further, as with a symmetric matrix, a normal matrix can be diagonalized as follows:
\begin{equation*}
\mat{W}=\Re(\mat{W}')=\Re(\mat{O}\diag(\mat{w}')\mat{O}^{*}).
\end{equation*}
Here, $\mat{O}\in\Cset^{n\times n}$ is a unitary matrix, $\mat{O}^{*}$ is the conjugate transpose of $\mat{O}$ and $\mat{w}'\in\Cset^{n}$ is the complex vector which consists of the eigenvalues of $\mat{W}'$. The bilinear transformation whose weight matrix $\mat{W}$ is replaced with this, can be transformed into (\ref{eq:prop2}). The unitary matrix $\mat{O}$ is a bijective function, so the input vectors $\mat{v}_i$, $\mat{v}_j$ are learned as their one-to-one correspondent $\mathbf{v}^{\prime}_{i}=\mathbf{O}\transpose \mathbf{v}_i$, $\mathbf{v}^{\prime}_{j} = \mathbf{O}\transpose \mathbf{v}_j$, assuming that the unitary matrix $\mathbf{O}$ is learned implicitly.
Note that to simultaneously diagonalize the normal matrices whose real parts are the weight matrices $\mat{U}^{[1]},\mat{U}^{[2]},\dots,\mat{U}^{[L]}$ in the scoring functions for labels, we have to assume that they form a commuting family as with the discussion in \ref{sim-diag}.
\section{Related work}
\label{sect:related_works}
\subsection*{Dependency Parsing}

In recent years, various graph-based parsers with attention mechanisms have been proposed. 

\cite{mlp-attention} incorporated the attention mechanism used in machine translation~\citep{attention} into their graph-based parser. Their model receives vectors which are made by concatenating LSTM outputs corresponding to each word and its head candidates. Similarly, \cite{bilinear-attention} proposed a graph-based parser where they substitute the MLP-based classifier in~\citep{mlp-attention} with the bilinear one in their multi-task neural model, although they still use the MLP-based one in prediction of labels. Accordingly,~\cite{deepbiaffine} modified the model by~\cite{mlp-attention} using a biaffine classifier instead of an MLP-based one which enables the model to express not only the probability of a word receiving a particular word as dependent but also the prior probability of a word being a head. 

Likewise, in transition-based parsing literature, the state-of-the-art parser on the English Penn TreeBank by~\cite{stackpointer} uses an attention mechanism based on a biaffne classifier which calculates the probability distribution of the next word which comes into the stack at each time step, with LSTM outputs corresponding to each word in the input sentence. The models proposed in this paper can be incorporated into these models.

\subsection*{Parameter Reduction in Neural Networks}
Recently, numerous methods toward parameter reduction of neural networks have been proposed.

As a similar approach to proposed methods, there is a method where a projection matrix is decomposed into smaller matrices by lower-rank approximation~\citep{lu:16}. In addition, \cite{ishihara:18} introduced eigenvector decomposition into neural tensor networks ~\citep{ntn} and analyzed the effects of parameter reduction. Although the paper~\citep{ishihara:18} is similar to the present paper in that their methods address parameter reduction in the bilinear term, our work is different in that we apply it to deep biaffine parser.

There are some methods to reduce parameters in a projection matrix by sharing them. \cite{yu:15} used circulant matrices in the fully connected layers. Our models are different from theirs in that we use circulant matrices in the bilinear terms. ~\cite{chen:2015} perform parameter reduction with a hash kernel and ~\cite{vikas:15} with special matrices like toeplitz matrices. While these can be also used for bilinear terms, the methods based on real diagonalization and circulant matrices are superior to them in computational efficiency.

~\cite{hinton:15} proposed a model called distillation and were able to train a model which was more compact than the original one. However, it needs a lot of time for training since it needs to be trained again for distillation. ~\cite{itay:16} achieved a significant reduction of parameters by the quantization, but the reported accuracy is inferior to the original model. Theoretically, these methods can be combined with the proposed methods.

\section{Experiments}
\label{sec:experiments}

\begin{table*}[t!]
\centering
\begin{tabular}{lcccccc}
\toprule
\bf Treebank & \multicolumn{2}{c}{\bf Baseline} & \multicolumn{2}{c}{\bf Symmetry Matrix} &     \multicolumn{2}{c}{\bf Circulant Matrix} \\
\bf             & \bf UAS          & \bf LAS  & \bf UAS      & \bf LAS  & \bf UAS     & \bf LAS \\\cmidrule(lr){1-7}
UD\_Czech & 93.72 & 91.89 & 93.45(-0.27) & 91.50(-0.39) & \bf 93.87(+0.15) & \bf 92.01(+0.12) \\
UD\_German & 87.57 & 84.27 & 87.21(-0.36) & 83.91(-0.36) & \bf 87.61(+0.04) & \bf 84.39(+0.12) \\
UD\_English & \bf 91.05 & \bf 89.42 & 90.95(-0.1) & 89.22(-0.2) & 91.04(-0.01)& 89.31(-0.11) \\
UD\_Chinese & 87.67 & 85.58 & 87.55(-0.12) & 85.41(-0.17) & \bf 87.96(+0.29) & \bf 85.65(+0.07) \\
UD\_Slovenian-SST & 75.63 & 69.53 & 74.94(-0.69) & 68.58(-0.95) & \bf 75.90(+0.27) & \bf 70.24(+0.71) \\
UD\_Latin & 70.90 & 64.53 & 70.27(-0.63) & 63.18(-1.35) & \bf 72.38(+1.48) & \bf 66.00(+1.47)\\
UD\_French-ParTUT & 91.82 & 89.78 & \bf 92.09(+0.27) & 89.94(+0.16) & 91.94(+0.12) & \bf 90.09(+0.31) \\
UD\_Galician-TreeGal & 80.10 & 74.77 & 80.05(-0.05) & 74.78 (+0.01) & \bf 80.24(+0.14) & \bf 75.68(+0.91) \\

\bottomrule
\end{tabular}
\caption{Main results on CoNLL 2017 dataset.}
\label{results-ud}
\end{table*}

\begin{table*}[t!]
\centering
\begin{tabular}{lcccccc}
\toprule
\bf Reduced Samples & \multicolumn{2}{c}{\bf Baseline} & \multicolumn{2}{c}{\bf Symmetry Matrix} &     \multicolumn{2}{c}{\bf Circulant Matrix} \\
\bf             & \bf UAS          & \bf LAS  & \bf UAS      & \bf LAS  & \bf UAS     & \bf LAS \\\cmidrule(lr){1-7}
0 / 4 & \bf 91.05 & \bf 89.42 & 90.95(-0.1) & 89.22(-0.2) & 91.04(-0.01) & 89.31(-0.11) \\
1 / 4 & \bf 90.32 & \bf 88.57 & 90.05(-0.27) & 88.30(-0.27) & 90.29(-0.03) & 88.49(-0.08) \\
2 / 4 & 88.98 & 87.08 & \bf 89.15(+0.17) & \bf 87.17(+0.09) & 88.72(-0.26) & 86.63(-0.45) \\
3 / 4 & 87.24 & 85.08 & 87.27(+0.03) & 85.06(-0.02) & \bf 87.59(+0.35) & \bf 85.38(+0.3) \\
\bottomrule
\end{tabular}
\caption{Results in UD\_English with fewer training samples.}
\label{results-reduced}
\end{table*}

\begin{table*}[t!]
\centering
\begin{tabular}{lcccccc}
\toprule

\bf Treebank & \bf UAS & \bf LAS \\ \cmidrule(lr){1-3}

UD\_Slovenian-SST & 74.98(-0.65) & 69.07(-0.46) \\
UD\_Latin & 71.96(+1.06) & 65.68(+1.15) \\
UD\_French-ParTUT &  92.17(+0.35) & 90.44(+0.66) \\
UD\_Galician-TreeGal & 79.68(-0.42) & 74.82(+0.05) \\

\bottomrule
\end{tabular}
\caption{Baseline model with reduced dimensions.
The numbers in parentheses are the differences from the baseline model with full dimensions.}
\label{results-halfdim}
\end{table*}

\subsection{Dataset and Implementation}

We compared the models described above on several languages in the CoNLL 2017 shared task for Universal Dependency Parsing dataset. We chose four languages which have relatively abundant training examples: UD\_Chinese, UD\_Czech, UD\_English and UD\_German. And we also selected four languages which have fewer training examples: UD\_French-ParTUT, UD\_Galician-TreeGal, UD\_Latin and UD\_Slovenian-SST. 

As a baseline model, we used the dependency parser by Timothy Dozat\footnote{https://github.com/tdozat/Parser-v2} which achieved the highest accuracy on the shared task. The structures of the proposed models are based on the baseline model; we changed only the classifier part.

We only modified two hyper-parameters: we used no pretrained embeddings and initialized word embeddings with a uniform distribution. These settings remain the same throughout all experiments unless otherwise stated. 

We use gold word segmentation and gold POS tags while word segmentation and POS tagging are included in the shared task. We excluded these two tasks because the objective of this research is to show the effects of the proposed methods on biaffine classifiers which are not used for those tasks.

\subsection{Results}
We show the results of the baseline and two proposed methods in Table~\ref{results-ud}. The method based on circulant matrices outperformed the others on almost all languages except for English where the baseline model achieved the best accuracy and French-ParTUT where the method based on symmetric matrix did so in UAS. Interestingly, the method based on symmetric matrices underperformed the baseline on most languages. This might be because of the restricted expressiveness of a symmetric weight matrix in comparison to a circulant one especially in that the former is not appropriate for expressing asymmetric relations.

\begin{table}[t!] 
\centering
\begin{tabular}{lrrr}
\toprule
& \bf \shortstack{Number of \\ Parameters} & \bf Percentage\\ \cmidrule(lr){1-3}
Character LSTM & 241200 & 7.64\% \\
Bidirectional LSTM & 1927200 & 61.03\% \\
Arc MLP & 320800 & 10.16\% \\
Label MLP & 80200 & 2.54\% \\
Arc Classifier & 160400 & 5.08\% \\
Label Classifier & 377437 & 11.95\% \\
Others & 50400 & 1.60\% \\ \hline
TOTAL & 3157637 & 100\% \\
\bottomrule
\end{tabular}
\caption{\label{results} Percentage of baseline model parameters accounted for by each component.}
\label{table:ratio}
\end{table}

\begin{table*}[t!]
\centering
\begin{tabular}{lrrr}
\toprule
\bf  & \bf Baseline &  \bf Symmetry Matrix & \bf Circulant Matrix \\ \cmidrule(lr){1-4}
Arc Classifier & 160400& 1200& 1600\\
Label Classifier & 377437& 11100& 14800\\
Sum with shared parts & 3157637 & 2632100 & 2636200 \\ \cmidrule(lr){1-4}
Difference from the baseline & 0.0\%& \bf -16.64\%& \bf -16.51\%\\
\bottomrule
\end{tabular}
\caption{\label{results} Comparison of parameter sizes.}
 \label{table:comparison}
\end{table*}

\section{Analysis}

In this section, we examine the robustness to overfitting of the proposed methods.

\subsection{Relaxation of Overfitting}

First, to further examine how numbers of parameters affect the models, we conducted experiments on UD\_English treebank reducing the number of training examples by a quarter at a time.

The results of this experiment are shown in Table \ref{results-reduced}. Our methods performed better on smaller datasets where the number of training examples are less than or equal to a half of the original number of examples. This indicates that our methods not only became robust to overfitting through parameter reduction but also achieved high generalizability. 

Second, we ran the baseline model with classifier dimensions reduced from 400 to 200 for the arc classifier and 100 to 50 for the label classifier and compared it with the proposed methods on languages from small treebanks. As shown in Table \ref{results-halfdim}, simply reducing the number of dimensions hindered the accuracies in some languages while the method based on circulant matrix consistently outperformed the baseline model with the original number of dimensions in all of these languages from small treebanks as shown in Table \ref{results-halfdim}. This result indicates the effectiveness of the proposed method on smaller datasets.

\subsection{Parameter Reduction}
Table \ref{table:ratio} shows the proportion of the total parameters which each part of the baseline model account for. While LSTMs occupy the largest portion of the model, the second largest part is the classifiers which account for about 17\% of the parameters. Table \ref{table:comparison} indicates that the proposed methods were able to reduce the number of parameters by more than 16\%.

\subsection{Parsing Speed}
To test the parsing speed, we used an NVidia GTX1080 GPU and parsed the test dataset of UD\_English.  As mentioned in Section~\ref{sec:proposed}, both proposed methods are superior to the baseline model in terms of time complexity. Actually, the methods based on symmetry  took 13.86 seconds, circularity 15.06 and the baseline 17.76 seconds. These results are in accordance with theoretical time complexity.

\section{Conclusion}
In this paper, we reduced the number of parameters in the weight matrices in biaffine classifiers based on the assumption of symmetry or circularity and examined their effects on the CoNLL 2017 shared task for Universal Dependency Parsing dataset. As a result, the method based on circulant matrices outperformed the baseline model in most of languages with about 16\% parameter reduction. As future work, the L1 regularization method for CompLex \citep{Trouillon:2016:CES:3045390.3045609} proposed by \citep{DBLP:conf/aaai/ManabeHS18}  may be integrated into our methods to further reduce the number of parameters in the bilinear function. The script for the bilinear functions proposed here is provided on the GitHub page below\footnote{https://github.com/TomokiMatsuno/PACLIC32/blob/master/my\_linalg.py}.

\section{Acknowledgments} 
We are grateful to Michael Wentao Li for proofreading the present paper. We thank the anonymous reviewers. This work was supported by JSPS KAKENHI Grant Number JP18K11457 and JST CREST Grant Number JPMJCR1513, Japan.











\bibliographystyle{plainnat}
{\footnotesize
\bibliography{reference}
}

\end{document}